\documentclass{article}

\usepackage[preprint]{neurips_2026}

\usepackage[utf8]{inputenc}
\usepackage[T1]{fontenc}
\usepackage{hyperref}
\usepackage{url}
\usepackage{booktabs}
\usepackage{amsfonts}
\usepackage{amsmath}
\usepackage{nicefrac}
\usepackage{microtype}
\usepackage{xcolor}
\usepackage{graphicx}
\usepackage{array}

\graphicspath{{figures/}}

\title{The Metacognitive Monitoring Battery:\\A Cross-Domain Benchmark for LLM Self-Monitoring}

\author{%
  Jon-Paul Cacioli \\
  Independent Researcher \\
  Melbourne, Australia \\
  ORCID: 0009-0000-7054-2014 \\
  \texttt{synthium@hotmail.com} \\
}

\begin{document}

\maketitle

\begin{abstract}
We introduce a cross-domain behavioural assay of monitoring-control coupling in LLMs, grounded in the Nelson and Narens (1990) metacognitive framework and applying human psychometric methodology to LLM evaluation. The battery comprises 524 items across six cognitive domains (learning, metacognitive calibration, social cognition, attention, executive function, prospective regulation), each grounded in an established experimental paradigm. Tasks T1--T5 were pre-registered on OSF prior to data collection; T6 was added as an exploratory extension. After every forced-choice response, dual probes adapted from Koriat and Goldsmith (1996) ask the model to KEEP or WITHDRAW its answer and to BET or decline. The critical metric is the withdraw delta: the difference in withdrawal rate between incorrect and correct items. Applied to 20 frontier LLMs (10{,}480 evaluations), the battery discriminates three profiles consistent with the Nelson--Narens architecture: blanket confidence, blanket withdrawal, and selective sensitivity. Accuracy rank and metacognitive sensitivity rank are largely inverted. Retrospective monitoring and prospective regulation appear dissociable ($r = .17$, 95\% CI wide given $n=20$; exemplar-based evidence is the primary support). Scaling on metacognitive calibration is architecture-dependent: monotonically decreasing (Qwen), monotonically increasing (GPT-5.4), or flat (Gemma). Behavioural findings converge structurally with an independent Type-2 SDT approach, providing preliminary cross-method construct validity. All items, data, and code: \url{https://github.com/synthiumjp/metacognitive-monitoring-battery}.
\end{abstract}

\section{Introduction}

\subsection{The metacognition gap in LLM evaluation}

When a language model answers a factual question, two capacities determine reliability: its ability to produce a correct answer, and its ability to monitor whether that answer is correct. These are different problems requiring different interventions. A model that produces correct answers but cannot distinguish its correct from incorrect outputs is unreliable for selective prediction, human-AI collaboration, or autonomous decision-making. A model that produces fewer correct answers but accurately flags its errors is often the more useful system. Current evaluation practice does not make this distinction. No standardised measurement tool exists for quantifying this capacity in LLMs.

Standard benchmarks (MMLU, Hendrycks et al., 2021; HumanEval, Chen et al., 2021; BIG-Bench, Srivastava et al., 2023) report accuracy, F1, or pass@k. A model that answers correctly with 100\% confidence is indistinguishable from one that could tell you which specific answers to trust. These benchmarks measure object-level performance (what the model knows) without measuring meta-level monitoring (whether the model knows what it knows).

This gap is beginning to receive attention. Steyvers and Peters (2025) reviewed LLM metacognition using AUROC and the meta-$d'$ framework. Kadavath et al. (2022) showed that language models can discriminate questions they answer correctly from those they do not. Ackerman (2025) introduced two behavioural paradigms (Delegate Game and Second Chance Game) to evaluate LLMs' strategic deployment of internal confidence signals, finding evidence of limited, context-dependent metacognition. Dai (2026) applied meta-$d'$ to verbalized confidence ratings and showed that scale design (0--20 vs.\ the standard 0--100) substantially affects metacognitive sensitivity through round-number discretization. This work collectively establishes that LLMs exhibit something that functions like metacognitive monitoring. However, the LLM literature borrows cognitive science terminology loosely, without applying the formal frameworks that give these terms their precision. To our knowledge, no prior LLM benchmark has explicitly operationalised the Nelson and Narens (1990) monitoring-control architecture as a benchmark design principle.

\subsection{The Nelson--Narens monitoring-control framework}

Nelson and Narens (1990) proposed that cognitive systems operate at two levels: an object level performing tasks and a meta-level monitoring and controlling it. Two information flows connect them: monitoring carries accuracy information upward; control adjusts behaviour downward. The critical insight is that monitoring and control can dissociate. A system can monitor without controlling (receives accuracy information but does not adjust behaviour), control without monitoring (applies a blanket policy without discriminating correct from incorrect outputs), or exhibit coupled monitoring-control. This three-way distinction is the theoretical spine of the benchmark. Koriat and Goldsmith (1996) extended the framework experimentally: in their free-report paradigm, participants answer questions and then decide whether to volunteer or withhold each answer, with the key metric being how well volunteer/withhold decisions track actual accuracy. We adapt this paradigm for LLMs. A further Nelson--Narens distinction is between retrospective monitoring (after-the-fact accuracy evaluation) and prospective regulation (strategy adjustment before responding); Metcalfe and Kornell (2005) demonstrated these are dissociable in humans. Our battery tests both.

\subsection{The dual-probe methodology}

After each forced-choice response, T1--T5 items administer two probes: ``KEEP or WITHDRAW this answer?'' and ``BET or NO\_BET that your answer is correct?'' These produce four response-commitment states per item, analysed independently of task scoring. The critical metric is the withdraw delta: the difference in withdrawal rate between incorrect and correct items. A positive delta indicates the commitment decision discriminates correct from incorrect; a delta near zero indicates a blanket policy. T6 uses a different probe structure: before answering, models choose ANSWER\_DIRECTLY, REQUEST\_HINT, or DECLINE, operationalising prospective regulation. Where meta-$d'$ (Maniscalco \& Lau, 2012; Cacioli, 2026b) measures how well an internal signal discriminates correct from incorrect, the withdraw delta measures how well the model's overt commitment decision does --- methodologically independent measurements of the same construct.

\subsection{Present study}

Metacognitive efficiency is domain-specific in humans (Fleming et al., 2014; Rouault et al., 2018), so a single-domain test cannot characterise a system's metacognitive capacity. Our battery spans six cognitive domains, each grounded in a distinct experimental paradigm. Tasks T1--T5 each have their own OSF pre-registration and (where available) a companion arXiv paper; T6 was added as an exploratory hackathon-track extension. We evaluate 20 frontier LLMs (10{,}480 total evaluations). Our contributions are: (1) \textbf{Methodological} --- a standardised measurement tool grounded in formal theory (Nelson \& Narens, 1990; Koriat \& Goldsmith, 1996) with convergent validity against an independent Type-2 SDT approach (Cacioli, 2026b); (2) \textbf{Three-profile taxonomy} (blanket confidence, blanket withdrawal, selective sensitivity) consistent with Nelson--Narens coupling states, with profiles fragmenting across domains within individual models; (3) \textbf{Dissociation} between retrospective monitoring and prospective regulation: the two measures are correlated $r = .17$ (wide CI given $n=20$) and diverge strongly for individual models; (4) \textbf{Architecture-dependent scaling} ruling out a universal scaling law: on T2, sensitivity decreases monotonically with scale for Qwen, increases for GPT-5.4, and remains flat for Gemma. All T1--T5 analyses were pre-registered on OSF; T6 is reported as exploratory. All items, data, and code are publicly archived.

\section{Method}

\subsection{Benchmark design principles}

The battery was designed around four principles. First, each task is grounded in a specific cognitive science paradigm with an established empirical literature. Second, each task includes diagnostic conditions that separate genuine competence from surface heuristics. Third, the five pre-registered tasks (T1--T5) are each paired with an OSF pre-registration and (where available) an arXiv companion paper; T6 was added as an exploratory extension. Fourth, every item carries metacognitive probes enabling the cross-domain analysis that is the battery's primary contribution.

\begin{table}[t]
\centering
\caption{Battery overview. Six tasks, six cognitive domains, 524 total items.}
\label{tab:battery}
\small
\begin{tabular}{llllrrll}
\toprule
Task & Domain & Paradigm & & Items & Conditions & OSF & arXiv \\
\midrule
T1 & Learning & Overhypothesis & & 98 & 8 & osf.io/h8un6 & 2603.13696 \\
T2 & Metacognition & SDT calibration & & 90 & 4 & osf.io/qpk9a & 2603.14893 \\
T3 & Social cognition & Mutual exclusivity & & 116 & 9 & osf.io/zu7af & 2603.13696 \\
T4 & Attention & Biased competition & & 60 & 6 & osf.io/d975z & -- \\
T5 & Executive & Weber's Law & & 88 & 3 & osf.io/5r76n & 2603.20642 \\
T6$^{\dagger}$ & Prospective & Calibrated help-seeking & & 72 & 3 & -- & -- \\
\bottomrule
\end{tabular}

\vspace{2pt}\footnotesize
$^{\dagger}$T6 was developed within the Kaggle AGI Hackathon 2026 as an exploratory extension. All items, scoring rules, and analytical specifications are archived on OSF.
\end{table}

\subsection{Task descriptions}

Representative items for each task, including the full probe sequence, are provided in Appendix A.

\textbf{T1: Learning (98 items).} Nonce-word world testing second-order generalisation (Kemp et al., 2007). Eight conditions from first-order retrieval through adversarial foils. Companion paper: Cacioli (2026c).

\textbf{T2: Metacognition (90 items).} Signal Detection Theory framework (Green \& Swets, 1966). Four conditions: calibration (66 items), prospective monitoring (8 items), error detection (8 items), knowledge boundaries (8 items). Companion paper: Cacioli (2026a).

\textbf{T3: Social Cognition (116 items).} Nine conditions from basic mutual exclusivity (Markman \& Wachtel, 1988) through scalar implicature, false belief at three orders (Perner \& Wimmer, 1985), and irony. Companion paper: Cacioli (2026c).

\textbf{T4: Attention (60 items).} Biased competition framework (Desimone \& Duncan, 1995). Six conditions testing selective attention under competition.

\textbf{T5: Executive Functions (88 items).} Three conditions through magnitude processing (Diamond, 2013): format flexibility (20 items), inhibitory control (43 items), task switching (25 items). Ratio-graded following Weber's Law (Dehaene, 2003). Companion paper: Cacioli (2026d).

\textbf{T6: Prospective Regulation (72 items).} Before answering, models choose ANSWER\_DIRECTLY (full credit if correct, zero if wrong), REQUEST\_HINT (half credit with hint), or DECLINE (quarter credit). Operationalises Metcalfe and Kornell's (2005) calibrated help-seeking paradigm. Retrospective probes follow. T6 should be interpreted as an assay of prospective regulation under explicit payoff contingencies rather than a comprehensive measure of prospective metacognition. T6 was developed within the Kaggle AGI Hackathon 2026 as an exploratory extension to the pre-registered battery. All items, scoring rules, and analytical specifications are archived on OSF alongside the five pre-registered tasks.

\subsection{The probe methodology}

After every forced-choice answer on T1--T5, two probes are administered. On T6, the prospective path choice is recorded before the response; retrospective probes follow.

The primary metric for retrospective monitoring is the withdraw delta:
\[
\Delta_{\text{withdraw}} = P(\text{WITHDRAW} \mid \text{incorrect}) - P(\text{WITHDRAW} \mid \text{correct})
\]

The primary metric for prospective regulation is the ANSWER\_DIRECTLY rate and its relationship to accuracy. ANSWER\_DIRECTLY is the prospective analog of the KEEP decision: the choice to commit to an answer without external support. A model that answers directly on most items is exhibiting minimal prospective regulation; a model that varies path choice with difficulty is exhibiting the regulatory behaviour Metcalfe and Kornell (2005) observed in human study-time allocation. REQUEST\_HINT and DECLINE rates are reported in supplementary analyses but are not the primary metric because they can be driven by generalised risk aversion rather than by calibrated difficulty detection.

\subsection{Models}

Twenty frontier LLMs from six provider families were evaluated. Selection enabled within-family scaling comparisons: Qwen (80B, 235B, 480B), GPT-5.4 (nano, mini, 5.4), Gemma (1B, 12B, 27B), as well as reasoning variants (DeepSeek R1 vs V3.2; Qwen Think vs Instruct) and architectural diversity across providers: Zhipu AI (GLM-5), Google (Gemini 3 Flash, 2.5 Flash, 3.1 Pro, 2.5 Pro; Gemma 1B/12B/27B), Anthropic (Opus 4.6, Sonnet 4.6, Haiku 4.5), OpenAI (GPT-5.4, mini, nano), Alibaba (Qwen 3 80B Think/Instruct, 235B, Coder 480B), DeepSeek (V3.2, R1).

\subsection{Evaluation platform and procedure}

All evaluations used the Kaggle Benchmarks platform (kbench SDK). Each item was administered independently with no cross-item context. Scoring was task-specific: accuracy for T1, T3, T4, T5; confidence-accuracy alignment for T2; path-weighted accuracy for T6. Probe responses were recorded separately and analysed independently.

\subsection{Profile classification}

Each model was classified per track into one of three profiles:
\begin{itemize}
    \item \textbf{Blanket confidence:} KEEP rate $\geq$ 95\% regardless of accuracy.
    \item \textbf{Blanket withdrawal:} KEEP rate $\leq$ 10\% regardless of accuracy (T1--T5) or DECLINE $\geq$ 90\% (T6).
    \item \textbf{Selective sensitivity:} Withdraw delta $\geq$ +15\%.
\end{itemize}

These thresholds are operational conventions chosen for interpretability, not claims about natural clustering or theoretically privileged cutoffs. The 95\%/10\%/15\% values sit at a stability plateau: all 20 models receive identical classifications at these thresholds, while shifting by $\pm$5 percentage points changes 9--10 of 20. Robustness across a wider threshold range is reported in \S\ref{sec:psychometric}.

\subsection{Analysis plan}

Three hypotheses were pre-registered across the five OSF filings covering T1--T5:

\textbf{H1:} No single model dominates all six domains.

\textbf{H2:} Domain-specific failure profiles are model-specific.

In addition, we report descriptively on which item types discriminate among models (previously pre-registered as H3) and on the exploratory hypothesis tested via T6:

\textbf{H4 (exploratory):} Retrospective monitoring and prospective regulation are dissociable.

\section{Results}

\subsection{Overall performance}

Table~\ref{tab:accuracy} presents the task-specific score matrix. Overall scores ranged from 0.594 (Gemma 1B) to 0.943 (Gemini 3 Flash). H1 was supported: no single model achieved the highest score on all six tracks.

\begin{table}[t]
\centering
\caption{Task-specific scores for selected models. T1, T3, T4, T5: simple accuracy. T2: mean metacognitive score (confidence-accuracy alignment, 0--1). T6: mean path-weighted score. Full 20-model table at \url{https://kaggle.com/benchmarks/jonpaulcacioli/classical-minds-modern-machines}.}
\label{tab:accuracy}
\small
\begin{tabular}{lrrrrrrr}
\toprule
Model & T1 & T2 & T3 & T4 & T5 & T6 & Overall \\
\midrule
Gemini 3 Flash & .969 & .906 & .974 & .883 & .989 & \textbf{.939} & \textbf{.943} \\
GLM-5 & .939 & \textbf{.942} & .957 & .933 & .989 & .858 & .936 \\
Claude Opus 4.6 & .857 & .921 & .957 & .967 & \textbf{1.00} & .924 & .938 \\
Gemini 3.1 Pro & .990 & .913 & .948 & .850 & .989 & .910 & .933 \\
Claude Sonnet 4.6 & .867 & .928 & .948 & .933 & .977 & .910 & .927 \\
GPT-5.4 & .929 & .888 & .957 & .950 & .943 & .837 & .917 \\
Claude Haiku 4.5 & .867 & .834 & .922 & \textbf{.983} & .886 & .803 & .883 \\
GPT-5.4 mini & .867 & .812 & .914 & .967 & .864 & .837 & .877 \\
DeepSeek R1 & \textbf{1.00} & .489 & \textbf{1.00} & \textbf{1.00} & \textbf{1.00} & .253 & .790 \\
GPT-5.4 nano & .714 & .661 & .871 & .900 & .784 & .710 & .773 \\
Gemma 1B & .622 & .741 & .672 & .800 & .614 & .115 & .594 \\
\bottomrule
\end{tabular}
\end{table}

T6 has the widest spread (0.824 range). R1's scores on T2 (0.489) and T6 (0.253) appear as accuracy failure but reflect behavioural withdrawal rather than competence failure: R1 declines 98.6\% of T6 items and withdraws 91--99\% of T1--T5 items.

\subsection{Three metacognitive profiles}
\label{sec:profiles}

All values in this section are single-administration point estimates per model; profile assignments should be read as descriptive classifications based on these point estimates rather than as bounded by test-retest confidence intervals. The probe data revealed three profiles mapping onto the monitoring-control coupling states predicted by Nelson and Narens (1990).

\paragraph{Profile A: blanket confidence (consistent with monitoring without control).}
Five models exhibited blanket confidence as a stable profile across all six tracks: Gemini 3 Flash, Gemini 2.5 Flash, Gemini 2.5 Pro, Gemini 3.1 Pro, and Qwen 80B Think. These responded KEEP on $\geq$ 95\% of T1--T5 items and ANSWER\_DIRECTLY on $\geq$ 97\% of T6 items regardless of correctness. Withdraw delta was near zero on every track. Gemini 3.1 Pro is classified by its dominant profile (mean KEEP = 97.1\%, mean withdraw delta = $-1.0\%$). In the Nelson--Narens framework, this profile is consistent with monitoring without control. The meta-level does not adjust response commitment based on accuracy. Not all blanket confidence was stable: GLM-5 showed blanket confidence on T2 ($\Delta = 0.0\%$) but strong selective sensitivity on T3 ($+59.1\%$).

\paragraph{Profile B: blanket withdrawal (DeepSeek R1, anomalous).}
DeepSeek R1 exhibited withdrawal rates of 91--99\% across T1--T5 and declined 98.6\% of T6 items, while simultaneously achieving near-ceiling accuracy where answers were scored (100\% on T1, T3, T4, T5). R1 therefore answers correctly and then retracts the correct answer, which is not control-without-monitoring in the standard Nelson--Narens sense; we interpret the pattern as chain-of-thought--induced blanket caution in which the reasoning process surfaces sufficient uncertainty during deliberation that the model withdraws on principle. The pattern is not universal to CoT (Qwen 80B Think is blanket confidence) nor to the architecture (DeepSeek V3.2 without extended CoT shows 97.8\% KEEP on T2). R1 is the sole model in this profile; we treat it as an outlier in aggregate statistics and exclude it from both the accuracy and metacognition rankings in Table~\ref{tab:inverted}.

\paragraph{Profile C: selective sensitivity (consistent with coupled monitoring-control).}
Eight models showed mean withdraw deltas $\geq$ +15\% across T1--T5: Claude Sonnet 4.6 ($+39.4\%$), Claude Haiku 4.5 ($+31.6\%$), Qwen Coder 480B ($+31.2\%$), Qwen 80B Instruct ($+29.5\%$), GPT-5.4 ($+27.9\%$), Qwen 235B Instruct ($+23.1\%$), GPT-5.4 mini ($+23.0\%$), and Gemma 3 12B ($+21.0\%$). Claude Haiku 4.5 showed the strongest domain-specific sensitivity: $+51.7\%$ on T2 and $+56.2\%$ on T3.

\begin{figure}[t]
\centering
\includegraphics[width=\linewidth]{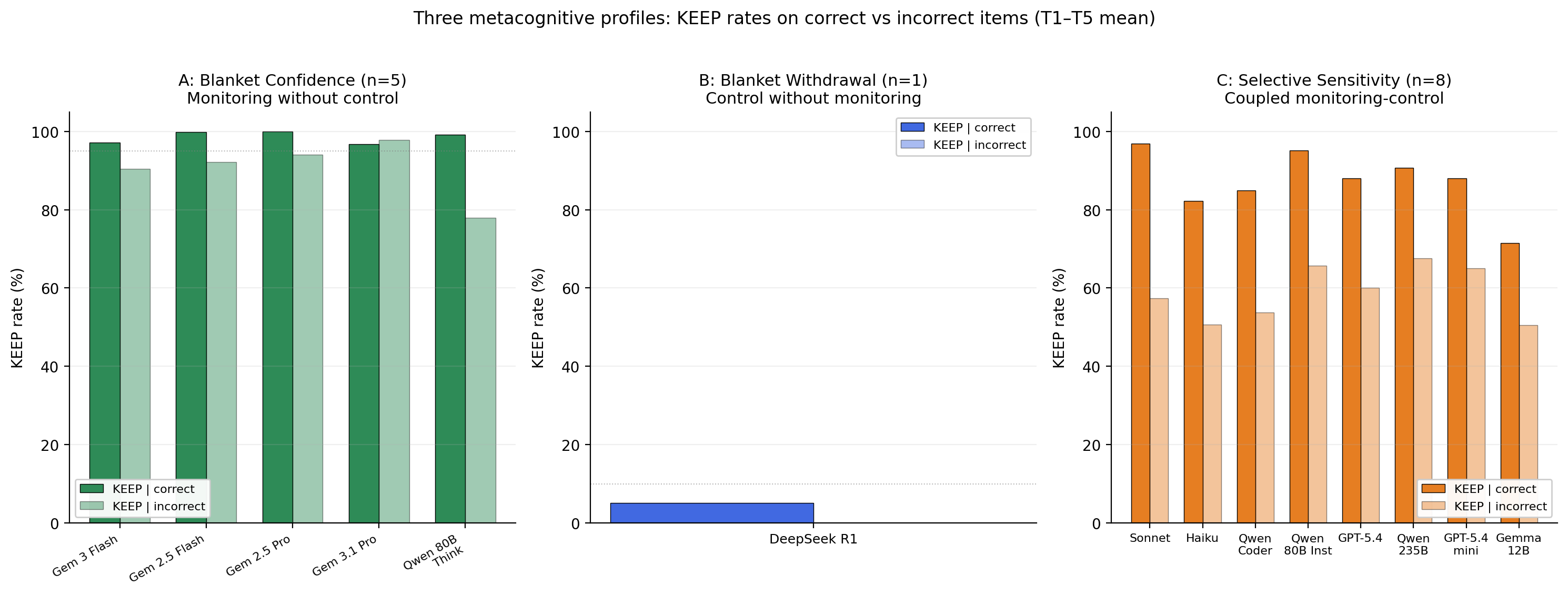}
\caption{KEEP rates on correct versus incorrect items for three metacognitive profiles. Points on the diagonal indicate identical KEEP rates regardless of correctness (no monitoring-control coupling). Points below the diagonal indicate selective withdrawal of incorrect items (coupled monitoring-control). \emph{Left:} Profile A models cluster on the diagonal at KEEP $\geq$ 95\%. \emph{Centre:} Profile B (DeepSeek R1) clusters at KEEP $\leq$ 10\% for both correct and incorrect items. \emph{Right:} Four representative Profile C models sit substantially below the diagonal.}
\label{fig:phenotypes}
\end{figure}

\subsection{The inverted leaderboard}

Accuracy rank and metacognitive sensitivity rank were approximately inverted. GLM-5 ranked 1st on T1--T5 accuracy (R1 excluded) but 9th on metacognitive sensitivity. Claude Haiku 4.5 ranked 13th on accuracy but 2nd on metacognitive sensitivity.

\begin{table}[t]
\centering
\caption{Inverted leaderboard (selected models). Acc rank: T1--T5 simple accuracy; Meta rank: mean W$\Delta$ (T1--T5). R1 excluded from both (\S3.2).}
\label{tab:inverted}
\small
\begin{tabular}{lrrr}
\toprule
Model & Acc Rank & Mean W$\Delta$ & Meta Rank \\
\midrule
Claude Sonnet 4.6 & 9 & $+39.4\%$ & 1 \\
Claude Haiku 4.5 & 13 & $+31.6\%$ & 2 \\
Qwen Coder 480B & 10 & $+31.2\%$ & 3 \\
Qwen 80B Inst & 12 & $+29.5\%$ & 4 \\
GPT-5.4 & 7 & $+27.9\%$ & 5 \\
Qwen 235B & 11 & $+23.1\%$ & 6 \\
GPT-5.4 mini & 14 & $+23.0\%$ & 7 \\
Gemma 3 12B & 17 & $+21.0\%$ & 8 \\
GLM-5 & 1 & $+14.6\%$ & 9 \\
GPT-5.4 nano & 18 & $+5.1\%$ & 17 \\
Qwen 80B Think & 4 & $+1.5\%$ & 18 \\
Gemini 3.1 Pro & 6 & $-1.0\%$ & 19 \\
\bottomrule
\end{tabular}
\end{table}

\begin{figure}[t]
\centering
\includegraphics[width=0.9\linewidth]{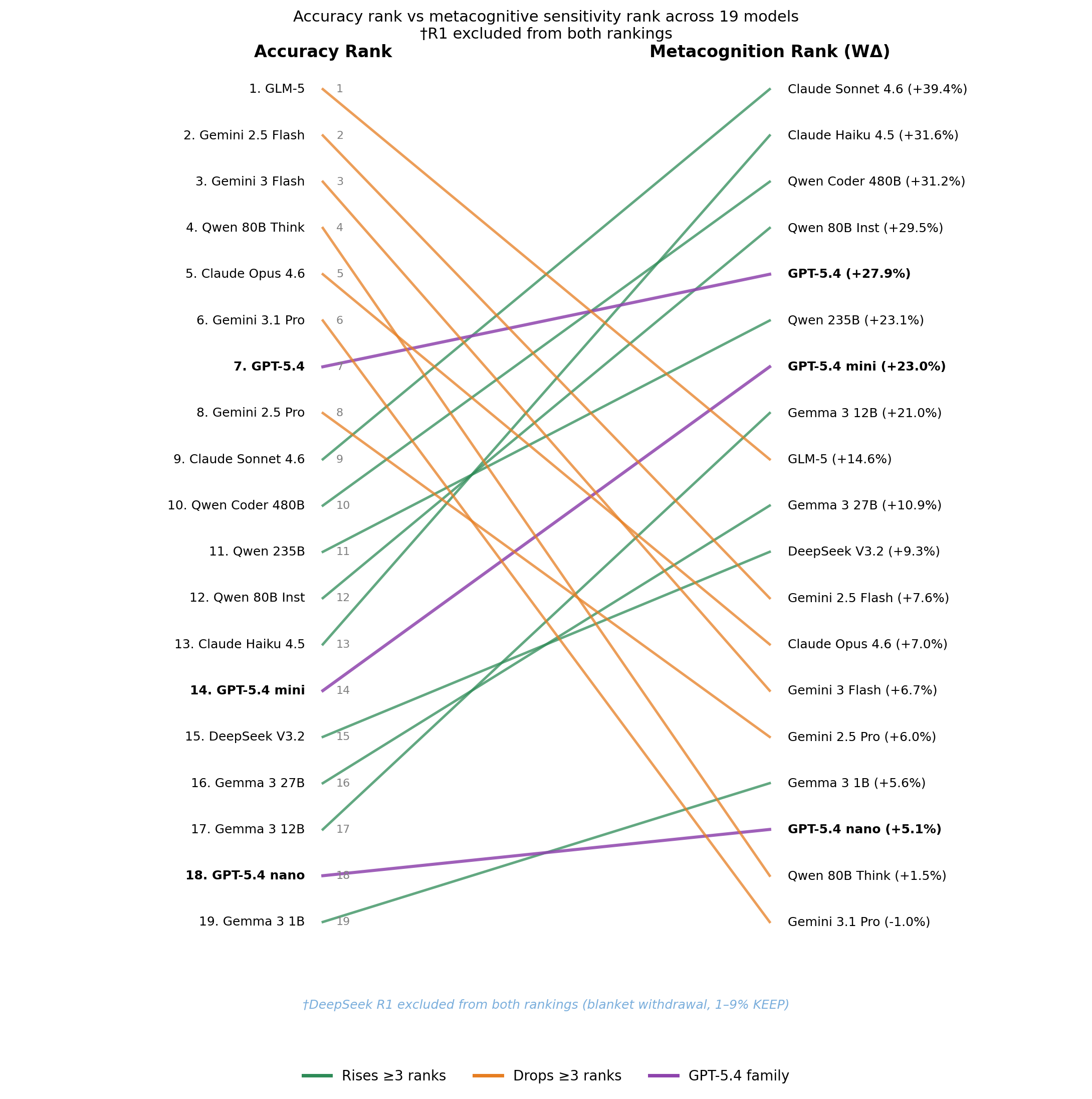}
\caption{Accuracy rank vs metacognitive sensitivity rank across 20 models. Green lines: models rising $\geq$ 3 ranks. Red/orange lines: models dropping $\geq$ 3 ranks. Purple lines: GPT-5.4 family. R1 excluded from both rankings.}
\label{fig:slope}
\end{figure}

\subsection{Retrospective--prospective dissociation}

Across 20 models, retrospective delta and T6 direct rate were weakly correlated (Pearson $r = .17$, $\rho = -.14$; 95\% Fisher-$z$ CI $[-.29, +.57]$). With $n=20$ the study is underpowered to detect moderate correlations (power for $r=.5$: 0.62; $r=.3$: 0.25), so the correlation alone does not establish independence. The strongest evidence for dissociability is the pattern of individual exemplars in which the two measures pull opposite ways:
\begin{itemize}
\setlength{\itemsep}{0pt}
    \item \textbf{Sonnet and GPT-5.4 (monitor, do not regulate).} Sonnet: highest retrospective sensitivity ($+39.4\%$) but 98.6\% ANSWER\_DIRECTLY. GPT-5.4: $+27.9\%$ and 98.6\% direct.
    \item \textbf{Haiku (monitors and regulates).} $+31.6\%$ retrospective and 84.7\% ANSWER\_DIRECTLY / 13.9\% REQUEST\_HINT; the only top model showing both.
    \item \textbf{Gemma 27B (regulates without monitoring well).} Retrospective $+10.9\%$; 36.1\% ANSWER\_DIRECTLY.
\end{itemize}

\subsection{Domain-specific metacognitive profiles}

Eight of 20 models showed stable profiles across all six tracks (5 Profile A, 1 Profile B, 2 Profile C). The remaining 12 showed domain-dependent profiles. The most dramatic case: Claude Sonnet 4.6 ranged from $+8.3\%$ on T1 to $+93.0\%$ on T5, a span of 85 percentage points within a single model. GLM-5 shifted from $0.0\%$ on T2 to $+59.1\%$ on T3. H2 was supported. These stability counts are based on single-administration point estimates per (model, track) and may partly reflect measurement noise; test-retest reliability is flagged as future work (\S\ref{sec:limitations}). Per-track profiles for all 20 models are included in the repository.

\subsection{Discriminating items}

Three item types caught the largest number of models. On T1's six adversarial foils, only 3 of 20 models answered any correctly; the remaining 17 scored 0\% while responding KEEP+BET on every item. On T2's knowledge-boundary items, every model scored 50--75\%, with no model reliably distinguishing its own knowledge from its own ignorance. On T3's scalar implicature items, most frontier models scored 50--62\%; only Gemini 3 Flash (88\%) and R1 (100\%) reliably handled cancellation.

\subsection{Architecture-dependent scaling}

We report scaling on T2 (metacognitive calibration), the track explicitly designed to measure metacognitive monitoring via SDT. The Qwen family showed T2 withdraw deltas decreasing monotonically with scale: 80B Instruct ($+30.1\%$) to 235B ($+21.8\%$) to Coder 480B ($+11.2\%$). The GPT-5.4 family showed monotonic increase: nano ($+17.1\%$) to mini ($+22.4\%$) to 5.4 ($+27.7\%$). The Gemma family was flat: 1B ($+18.3\%$), 12B ($+18.3\%$), 27B ($+16.7\%$). Three families, three trajectories on the same metric. These patterns argue against assuming a universal monotonic scaling relation for metacognitive sensitivity. Mean withdraw delta across T1--T5 produces different patterns for some families, suggesting that scaling effects on metacognition are domain-dependent as well as architecture-dependent.

\begin{figure}[tb]
\centering
\includegraphics[width=0.78\linewidth]{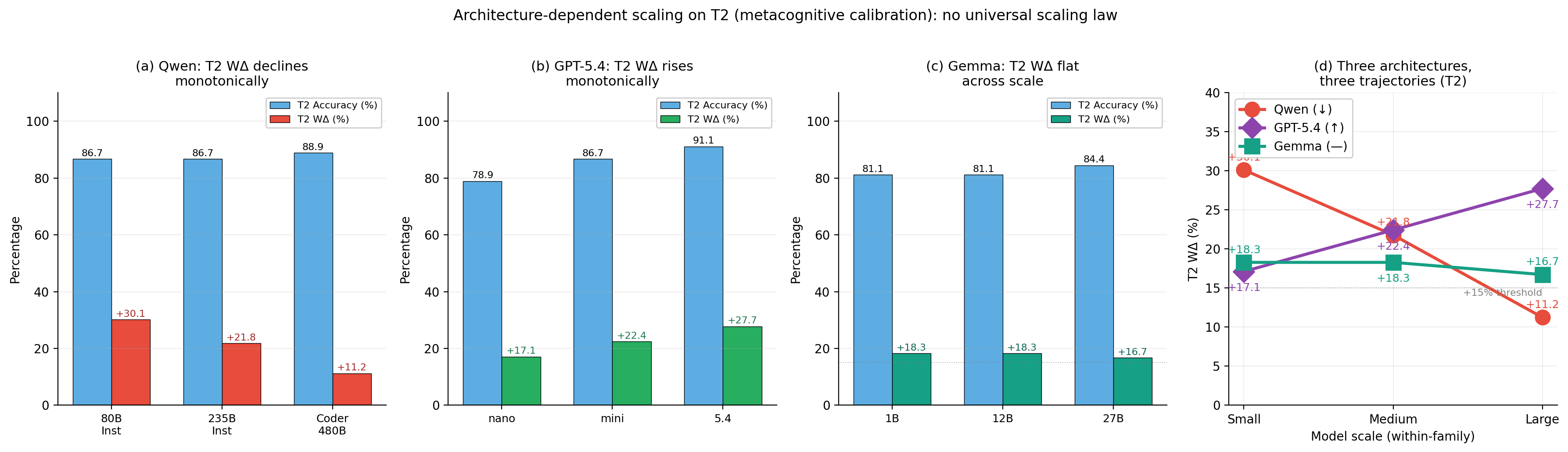}
\caption{Scaling and metacognition on T2 across three model families. (a) Qwen: T2 accuracy stable while T2 W$\Delta$ declines monotonically with scale. (b) GPT-5.4: T2 W$\Delta$ increases monotonically with scale. (c) Gemma: T2 W$\Delta$ flat ($\sim +17$--$18\%$). (d) Three families, three trajectories on the same metric.}
\label{fig:scaling}
\end{figure}

\subsection{Convergent validity with Type-2 SDT}

Cacioli (2026b) applied Type-2 Signal Detection Theory to four open-weight LLMs (Qwen-7B, Llama-3.1-8B, Mistral-7B, Gemma-2-9B) across 224{,}000 factual QA trials using token-level log-probabilities rather than behavioural probes. That study's AUROC2 and M-ratio rankings fully inverted relative to Type-1 accuracy --- the same structural pattern as our inverted leaderboard (\S3.3). Different method, different models, different items, same inverted-ranking structure: convergence consistent with the inversion reflecting a real property of current LLM metacognition rather than a method artifact (cf.\ Campbell \& Fiske, 1959). Same-item Type-2 SDT on the full battery requires logprob access to closed-source models and is planned as a separate study.

\subsection{Psychometric properties}
\label{sec:psychometric}

Cronbach's $\alpha$ across the six tracks was 0.54. A high $\alpha$ would indicate the six tracks measure the same latent construct; the Nelson--Narens framework and human metacognition literature (Fleming et al., 2014; Rouault et al., 2018) predict they should not, because metacognitive monitoring is partially domain-specific. The moderate $\alpha$ is thus consistent with theoretical prediction rather than a psychometric deficit. Split-half reliability (T1+T2+T4 vs T3+T5+T6) was $r = .51$, Spearman--Brown corrected 0.68. Cohen's $d$ separating Profile A ($n=5$) from Profile C ($n=8$) was 4.57 (95\% bootstrap CI $[3.65, 7.84]$, 10{,}000 resamples); because profiles are defined using withdraw delta this reflects descriptive separation rather than independent validation, but the lower bound is well above the conventional large-effect threshold. At the chosen thresholds all 20 models matched their reported classification; shifting by $\pm$5 points changes 9--10 of 20. Accuracy and mean withdraw delta are approximately independent ($r = .16$).

\section{Discussion}

\subsection{What the three profiles mean}

Nelson and Narens (1990) predicted three coupling states between monitoring and control; we observe behavioural profiles consistent with all three, used descriptively rather than as claims about discrete underlying categories. We do not claim these profiles reflect the same mechanisms as human metacognition. Profile C maps onto the normative adult pattern described by Koriat and Goldsmith (1996), against which impairments are measured. For systems using confidence for selective decisions (abstention, delegation, verification), Profile C models are likely most useful: Profile A cannot abstain informatively, Profile B abstains on everything. In deployment contexts with external verification, high-accuracy Profile A models may still be preferable; the battery augments rather than replaces accuracy metrics.

\subsection{Blanket confidence and the inverted leaderboard}

On T1's six adversarial foils, 17 of 20 models scored 0\% while responding KEEP+BET on every item: maximum confidence on maximum failure, stable across families and scales. Standard accuracy benchmarks assign no penalty for confident failure; the withdraw delta exposes it. More broadly, a model that responds KEEP on every item scores its full accuracy under the standard metric --- blanket endorsement is rewarded. Any post-training intervention that raises the endorsement rate without dropping accuracy will appear to improve performance. The three-profile taxonomy provides a way to detect this; we recommend reporting metacognitive metrics (withdraw delta, M-ratio) alongside accuracy. The present data do not distinguish three candidate mechanisms for the inversion: accuracy and monitoring are genuinely anti-correlated at the frontier; post-training that pushes accuracy up also decouples confidence from accuracy; or the probe format elicits different response policies across families for reasons unrelated to monitoring. Same-item Type-2 SDT (\S3.8) would help distinguish these.

\subsection{Two dissociations}

Retrospective monitoring and prospective regulation appear separable. The correlation ($r = .17$, 95\% CI $[-.29, +.57]$) does not establish independence with $n = 20$; the dissociation pattern is strongest in the exemplars: Sonnet and GPT-5.4 monitor without regulating, Gemma 27B regulates without monitoring, Haiku does both. On T2, three architectures trace three distinct scaling relationships (Qwen: decreasing; GPT-5.4: increasing; Gemma: flat), ruling out a universal monotonic scaling relation. Differences in post-training regimes --- particularly confidence-related reward signals during RLHF --- may drive the divergent trajectories, though the current data cannot adjudicate between training-based and architecture-based explanations.

\subsection{Domain specificity and programme-level design}

The fragmentation parallels the human neuroscience literature (Fleming et al., 2014; Rouault et al., 2018): no single-domain test characterises monitoring capacity. This battery is part of a multi-paper programme with pre-registered companion papers on overhypothesis induction (Cacioli, 2026c), SDT calibration (Cacioli, 2026a), and Weber's Law (Cacioli, 2026d). The present paper stands alone.

\subsection{Limitations and future directions}
\label{sec:limitations}

Per-condition cell sizes are small on some tracks (T4: 10 items per condition). The binary KEEP/WITHDRAW probe may measure response style as much as monitoring; convergent validity with Type-2 SDT and moderate internal consistency partially address this. Each item was administered once per model, so profile assignments are single-administration point estimates. Planned extensions: graded confidence probes enabling direct meta-$d'$ computation, test-retest reliability, hierarchical Bayesian estimation (HMeta-$d'$; Fleming, 2017), and human-LLM comparisons.

\section*{Open science and AI disclosure}

Data, code, and Croissant metadata: \url{https://github.com/synthiumjp/metacognitive-monitoring-battery}. All analyses reported. Claude (Anthropic) was used for pipeline implementation and manuscript preparation; scientific decisions were the author's.

\section*{References}

{\small
\setlength{\parindent}{-0.2in}
\setlength{\leftskip}{0.2in}
\setlength{\parskip}{4pt}

Ackerman, C. (2025). Evidence for limited metacognition in LLMs. \emph{arXiv:2509.21545}. Published as ICLR 2026 conference paper.

Cacioli, J. P. (2026a). LLMs as signal detectors. \emph{arXiv:2603.14893}.

Cacioli, J. P. (2026b). Do LLMs know what they know? \emph{arXiv:2603.25112}.

Cacioli, J. P. (2026c). Exemplar retrieval without overhypothesis induction in large language models. \emph{arXiv:2603.13696}.

Cacioli, J. P. (2026d). Weber's Law in transformer magnitude representations. \emph{arXiv:2603.20642}.

Campbell, D. T., \& Fiske, D. W. (1959). Convergent and discriminant validation by the multitrait-multimethod matrix. \emph{Psychological Bulletin, 56}, 81--105.

Chen, M., et al. (2021). Evaluating large language models trained on code. \emph{arXiv:2107.03374}.

Dai, Y. (2026). Rescaling confidence: What scale design reveals about LLM metacognition. \emph{arXiv:2603.09309}.

Dehaene, S. (2003). The neural basis of the Weber--Fechner law. \emph{Trends in Cognitive Sciences, 7}, 145--147.

Desimone, R., \& Duncan, J. (1995). Neural mechanisms of selective visual attention. \emph{Annual Review of Neuroscience, 18}, 193--222.

Diamond, A. (2013). Executive functions. \emph{Annual Review of Psychology, 64}, 135--168.

Fleming, S. M. (2017). HMeta-$d$. \emph{Neuroscience of Consciousness, 2017}(1), nix007.

Fleming, S. M., \& Lau, H. C. (2014). How to measure metacognition. \emph{Frontiers in Human Neuroscience, 8}, 443.

Fleming, S. M., Ryu, J., Golfinos, J. G., \& Blackmon, K. E. (2014). Domain-specific impairment in metacognitive accuracy following anterior prefrontal lesions. \emph{Brain, 137}, 2811--2822.

Green, D. M., \& Swets, J. A. (1966). \emph{Signal detection theory and psychophysics.} Wiley.

Hendrycks, D., et al. (2021). Measuring massive multitask language understanding. In \emph{Proceedings of ICLR}.

Kadavath, S., et al. (2022). Language models (mostly) know what they know. \emph{arXiv:2207.05221}.

Kemp, C., Perfors, A., \& Tenenbaum, J. B. (2007). Learning overhypotheses with hierarchical Bayesian models. \emph{Developmental Science, 10}, 307--321.

Koriat, A., \& Goldsmith, M. (1996). Monitoring and control processes in the strategic regulation of memory accuracy. \emph{Psychological Review, 103}, 490--517.

Maniscalco, B., \& Lau, H. (2012). A signal detection theoretic approach for estimating metacognitive sensitivity from confidence ratings. \emph{Consciousness and Cognition, 21}, 422--430.

Markman, E. M., \& Wachtel, G. F. (1988). Children's use of mutual exclusivity. \emph{Cognitive Psychology, 20}, 121--157.

Metcalfe, J., \& Kornell, N. (2005). A region of proximal learning model of study time allocation. \emph{Journal of Memory and Language, 52}, 463--477.

Nelson, T. O., \& Narens, L. (1990). Metamemory. In G. H. Bower (Ed.), \emph{The psychology of learning and motivation} (Vol.~26, pp.~125--173). Academic Press.

Perner, J., \& Wimmer, H. (1985). Attribution of second-order beliefs. \emph{Journal of Experimental Child Psychology, 39}, 437--471.

Rouault, M., et al. (2018). Human metacognition across domains. \emph{Personality Neuroscience, 1}, e17.

Srivastava, A., et al. (2023). Beyond the imitation game. \emph{Transactions on Machine Learning Research}.

Steyvers, M., \& Peters, M. A. K. (2025). Metacognition and uncertainty communication in humans and large language models. \emph{Current Directions in Psychological Science}. Advance online publication.
}

\appendix

\section{Representative items}

\textbf{T1 (Learning).} ``In Domain A: a `blicket' is round, red, smooth; a `dax' is square, blue, rough; a `wug' is triangular, green, soft. A new `fep' is star-shaped, yellow, bumpy. Which is more likely also a `fep': (A) star-shaped, orange, fuzzy, or (B) circular, yellow, bumpy? [A.] KEEP or WITHDRAW? BET or NO\_BET?''

\textbf{T2 (Metacognition).} ``Claim: `The speed of light is approximately $3 \times 10^8$ m/s.' TRUE or FALSE? Confidence 1--5. [TRUE, 5.] KEEP or WITHDRAW? BET or NO\_BET?''

\textbf{T3 (Social Cognition).} ``'Some students passed.' All 30 passed. Was the statement (A) technically true, or (B) false? [A.] KEEP or WITHDRAW? BET or NO\_BET?''

\textbf{T4 (Attention).} ``Ambiguous passage with preceding context priming Interpretation 1. Which: (A) or (B)? KEEP or WITHDRAW? BET or NO\_BET?''

\textbf{T5 (Executive).} ``Which is larger: (A) 0.000999, or (B) 0.00100? [B.] KEEP or WITHDRAW? BET or NO\_BET?''

\textbf{T6 (Prospective).} ``Choose: (A) ANSWER\_DIRECTLY, (B) REQUEST\_HINT, or (C) DECLINE. Question: `What is the chemical formula for water?' [Model chooses path, answers if A/B.] KEEP or WITHDRAW? BET or NO\_BET?''

%%% Paper checklist omitted from arXiv preprint; available in the NeurIPS submission version.
%\input{checklist.tex}

\end{document}